\documentclass[10pt,journal,compsoc]{IEEEtran}

\ifCLASSOPTIONcompsoc
\usepackage[nocompress]{cite}
\else
\usepackage{cite}
\fi
\usepackage{cite}
\usepackage{hyperref}
\usepackage{graphicx}
\usepackage{subfigure}
\usepackage{amsmath}
\usepackage{booktabs}
\usepackage{multirow}
\usepackage{color}
\usepackage{ragged2e}
\usepackage{caption}
\usepackage{xspace}
\usepackage{etoolbox} 
\usepackage{amsfonts,amssymb}

\makeatletter
\DeclareRobustCommand\onedot{\futurelet\@let@token\@onedot}
\def\@onedot{\ifx\@let@token.\else.\null\fi\xspace}

\def\ie{\emph{i.e}\onedot} 
 
\def\etc{\emph{etc}\onedot} 
 
\def\etal{\emph{et al}\onedot}
\makeatother

\ifCLASSINFOpdf

\else

\fi

\begin{document}
	
	\title{End-to-end Alternating Optimization for Blind Super Resolution}
	
	\author{
		Zhengxiong~Luo,
		Yan~Huang,
		Shang~Li,
		Liang~Wang,~\IEEEmembership{Fellow,~IEEE,}
		Tieniu~Tan,~\IEEEmembership{Fellow,~IEEE}
		\IEEEcompsocitemizethanks{
			\IEEEcompsocthanksitem Z. Luo is with the University of Chinese Academy of Sciences (UCAS), Beijing 100049, P.R. China, the Center for Research on Intelligent Perception and Computing (CRIPAC), National Laboratory of Pattern Recognition (NLPR), Institute of Automation, Chinese Academy of Sciences (CASIA), Beijing 100190, China. E-mail: zhengxiong.luo@cripac.ia.ac.cn. 
			\IEEEcompsocthanksitem Y. Huang is with the Center for Research on Intelligent Perception and Computing (CRIPAC), National Laboratory of Pattern Recognition (NLPR), Institute of Automation, Chinese Academy of Sciences (CASIA), Beijing 100190, China, and with the University of Chinese Academy of Sciences (UCAS), Beijing 100049, P.R. China. E-mail: yhuang@nlpr.ia.ac.cn. 
			\IEEEcompsocthanksitem S. Li is with the University of Chinese Academy of Sciences (UCAS), Beijing 100049, P.R. China and Institute of Automation, Chinese Academy of Sciences (CASIA), Beijing 100190, China. E-mail: lishang2018@ia.ac.cn.
			\IEEEcompsocthanksitem L. Wang and T. Tan are with the Center for Research on Intelligent Perception and Computing (CRIPAC), National Laboratory of Pattern Recognition (NLPR), Center for Excellence in Brain Science and Intelligence Technology (CEBSIT), Institute of Automation, Chinese Academy of Sciences (CASIA), Beijing 100190, China, and with the University of Chinese Academy of Sciences (UCAS), Beijing 100049, P.R. China. E-mail: {wangliang, tnt}@nlpr.ia.ac.cn.
		}
	}
	
	\IEEEtitleabstractindextext{%
		\begin{abstract}
			\justifying
			Previous methods decompose the blind super-resolution (SR) problem into two sequential steps: \textit{i}) estimating the blur kernel from given low-resolution (LR) image and \textit{ii}) restoring the SR image based on the estimated kernel. This two-step solution involves two independently trained models, which may not be well compatible with each other. A small estimation error of the first step could cause a severe performance drop of the second one. While on the other hand, the first step can only utilize limited information from the LR image, which makes it difficult to predict a highly accurate blur kernel. Towards these issues, instead of considering these two steps separately, we adopt an alternating optimization algorithm, which can estimate the blur kernel and restore the SR image in a single model. Specifically, we design two convolutional neural modules, namely \textit{Restorer}  and \textit{Estimator}. \textit{Restorer} restores the SR image based on the predicted kernel, and \textit{Estimator}  estimates the blur kernel with the help of the restored SR image. We alternate these two modules repeatedly and unfold this process to form an end-to-end trainable network. In this way, \textit{Estimator} utilizes information from both LR and SR images, which makes the estimation of the blur kernel easier. More importantly, \textit{Restorer} is trained with the kernel estimated by \textit{Estimator}, instead of the ground-truth kernel, thus \textit{Restorer} could be more tolerant to the estimation error of \textit{Estimator}. Extensive experiments on synthetic datasets and real-world images show that our model can largely outperform state-of-the-art methods and produce more visually favorable results at a much higher speed. The source code is available at \url{https://github.com/greatlog/DAN.git}. 
		\end{abstract}
		
		}

	\maketitle

	\IEEEdisplaynontitleabstractindextext

	%
	\IEEEpeerreviewmaketitle
	
	\IEEEraisesectionheading{\section{Introduction}\label{sec:introduction}}
	
	\IEEEPARstart{S}{ingle} image super-resolution (SISR) aims to recover the high-resolution (HR) version of a given degraded low-resolution (LR) image. It has wide applications in video enhancement, medical imaging, as well as security and surveillance imaging. Mathematically, the degradation process can be expressed as
	\begin{equation}
		\mathbf{y} = (\mathbf{x}\otimes \mathbf{k})\downarrow_{s} + \mathbf{n} \label{downsample}
	\end{equation}
	where $\mathbf{x}$ is the original HR image, $\mathbf{y}$ is the degraded LR image, $\otimes$ denotes the two-dimensional convolution of $\mathbf{x}$ with blur kernel $\mathbf{k}$, $\mathbf{n}$ denotes Additive White Gaussian Noise (AWGN), and $\downarrow_{s}$ denotes the standard $s$-fold downsampler, which means keeping only the upper-left pixel for each distinct $s\times s$ patch~\cite{usr}. Then SISR refers to the process of recovering $\mathbf{x}$ from $\mathbf{y}$. It is a highly ill-posed problem due to this inverse property, and thus has always been a challenging task~\cite{baker2002limits}.
	
	During the past five years, deep neural networks (DNNs) have achieved remarkable results on SISR~\cite{anwar2020deep,survey}. But most of these methods \cite{imdn,idn} assume that the blur kernel is predefined as the kernel of bicubic interpolation. In this case, the SR task degenerates to find the inverse solution for bicubic downsampling. However, blur kernels in real applications are much more complicated. They are usually unknown and differ from image to image, as the blur kernels can be easily influenced by the camera intrinsic parameters, camera pose, \etc. Consequently, there is a domain gap between bicubically synthesized training samples and the real images. This domain gap will lead to a severe performance drop when these networks are applied to real applications~\cite{bridge}. Thus, more attention should be paid to SR in the context of unknown blur kernels, \ie blind SR. 
	
	In blind SR, there is one more undetermined variable, \ie the blur kernel $\mathbf{k}$, and the optimization also becomes much more difficult. To make this problem easier to be solved, previous methods~\cite{srmd,dpsr} usually decompose it into two sequential steps: \textit{i}) estimating the blur kernel from LR image and \textit{ii}) restoring the SR image based on estimated kernel. This two-step solution involves two independently trained models, thus they may be not well compatible with each other. Specifically, the model in the second step is usually trained with ground-truth kernels. While during testing, it is provided with kernel estimated in the first step. As a result, a small estimation error of the first step could cause a severe performance drop of the following one~\cite{ikc}. And on the other hand, the first step can only utilize limited information from the LR image, which makes it difficult to predict a highly accurate blur kernel. Consequently, although both models can perform well individually, the final result may be suboptimal when they are combined together. 
	
	Instead of considering these two steps separately, we adopt an alternating optimization algorithm, which can estimate the blur kernel $\mathbf{k}$ and restore the SR image $\mathbf{x}$ in the same model. In detail, we design two convolutional neural modules, namely \textit{Restorer} and \textit{Estimator}. \textit{Restorer} restores the SR image based on the blur kernel predicted by \textit{Estimator}, and the restored the SR image is further used to help \textit{Estimator} estimate a better blur kernel. Once the blur kernel is manually initialized, the two modules can well corporate with each other to form a closed loop, which can be iterated over and over. The iterating process is then unfolded to an end-to-end trainable network, which is called a deep alternating network (DAN). In this way, \textit{Estimator} can utilize information from both LR and the SR images, which makes the estimation of  the blur kernel easier. More importantly, \textit{Restorer} is trained with the kernel estimated by \textit{Estimator}, instead of ground-truth kernel. Thus during testing \textit{Restorer} could be more tolerant to the estimation error of \textit{Estimator}. Besides, the results of both modules could be substantially improved during the iterations, thus it is likely for our alternating optimization algorithm to get better final results than the direct two-step solutions. 
	
	We summarize our contributions into three points:
	\begin{itemize}
		\item [1.] We adopt an alternating optimization algorithm to estimate the blur kernel and restore the SR image for blind SR in a single network (DAN),  which helps the two modules to be well compatible with each other and so as to get better final results than the previous two-step solution. 
		\item [2.] We design two convolutional neural modules, which can be alternated repeatedly and then unfolded to form an end-to-end trainable network, without any pre/post-processing. It is easier to be trained and has a higher speed than the previous two-step solution. To the best of our knowledge, the proposed method is the first end-to-end network for blind SR.
		\item [3.] Extensive experiments on synthetic and real-world images show that our model can largely outperform state-of-the-art methods and produce more visually favorable results at a much higher speed.
	\end{itemize}
	
	A preliminary version of this work has been presented as a conference paper~\cite{dan}. In the current work, we incorporate additional contents in significant ways:
	\begin{itemize}
		\item[1.] We propose a dual-path conditional block (DPCB) to optimize the architectures of  both \textit{Estimator} and \textit{Restorer}~(Sec~\ref{basic_block}). Compared with the original conditional residual block (CRB), DPCB has its advantages: \textit{i}) DPCB can simultaneously explore deep features of both its basic and conditional inputs, while CRB only focuses on the basic one. It enables DPCB to model a deeper correlation between the two inputs and help improve the performance of \textit{Estimator} and \textit{Restorer}. \textit{ii}) The dual-path design in DPCB abandons the expansion and concatenation operations in CRB, which saves much computation. Experiments show that DPCB accelerates the whole network by 28\%. 
		\item[2.] In current version, \textit{Estimator} is supervised by the complete blur kernel, instead of the kernel in the reduced space as the conference version does. On the one hand, stronger supervision may help \textit{Estimator} to be better optimized. On the other hand,  it is easy for the complete predicted kernel to be used in other tasks, while the reduced kernel can only be used in the \textit{Restorer}. 
		\item[3.] We investigate more details and add considerable analysis to the initial version, such as visualization of the predicted kernel, ablation studies about the architectures of \textit{Restorer} and \textit{Estimator},  \etc.
	\end{itemize}  
	
	\section{Related Work}
	\subsection{Super Resolution for Bicubic Downsampling}
	Learning-based methods for SISR usually require a large number of paired HR and LR images as training samples. However, these paired samples are hard to obtain in the real world. Consequently, researchers manually synthesize LR images from HR images with predefined downsampling settings. The most popular setting is bicubic interpolation, \ie defining $\mathbf{k}$ in Equation~\ref{downsample} as the bicubic kernel. In this way, a large amount of paired samples can be easily synthesized, which helps boost the development of various deep-learning-based methods. Since the arising of SRCNN~\cite{srcnn}, various DNNs~\cite{survey,dbpn,meta_sr} have been proposed based on this setting. And most of them focus on optimizing the network architecture for SR. Strategies such as  post-upscaling~\cite{fsrcnn}, residual learning~\cite{vdsr}, and pixel-shuffle operation~\cite{pixel_shuffle}, have become the default choices for building an SR network. After the proposal of RCAN~\cite{rcan}, RRDB~\cite{esrgan} and SAN~\cite{san}, the performance in the context of bicubic downsampling even starts to get saturate on common benchmark datasets. 
	
	Despite that great achievements have been made for super-resolving bicubically downsampled images, it is still difficult for SR methods to get applied in real scenarios. Because the blur kernels for real images are usually unknown and differ from image to image,  and are much more complicated than the bicubic one. Consequently, due to the domain gap between real and synthesized data, methods designed for bicubically downsampled images will suffer serve performance drop in real applications~\cite{bridge,cycleSR}.  To address this issue, researchers begin to work on more challenging cases where degradations of test images are unknown, \ie blind super resolution.
	
	\subsection{Two-step Blind Super Resolution}~\label{non-blind}
	As indicated in Equation~\ref{downsample},  blind super resolution involves solving both the blur kernel $\mathbf{k} $ and SR image $\mathbf{x}$. Previous methods usually decompose it into two sequential steps and each step is an independent research field.
	
	\vspace{0.02\linewidth}\noindent\textbf{Kernel estimation.}
	The first step is estimating the blur kernel from the test image. As this is an ill-posed problem~\cite{levin,levin2011efficient}, some priors are usually needed to get it properly solved. In~\cite{nonpara}, a non-parametric method is used by utilizing the patch recurrence between the test image and its downscaled version. A similar idea is also adopted in~\cite{kernel_gan,gan_first}, but powered with neural networks and adversarial training~\cite{gan}. Another widely used prior is the extreme channel priors. In~\cite{pan,pan2016blind},  Pan \etal firstly propose the dark channel prior, \ie the dark channel in a natural image is usually sparse, which can be used for solving the blur kernel from a blurred image. In~\cite{extreme,cai2020dark}, the bright channel prior is further proposed and the idea is augmented to extreme channel priors. Although these manually set priors may help in some cases,  they may often be violated in applications. Consequently, as we will show in the experimental section~\ref{kernel_study}, the accuracy of estimated kernels is still limited. 
	
	\vspace{0.02\linewidth}\noindent\textbf{Super Resolution with given kernel.}
	The second step is super-resolving the SR image with the estimated kernel. This research field is also known as non-blind SR, in which methods are designed under the assumption that the ground-truth blur kernel is known. In~\cite{zssr_pre,zssr,mzsr}, the blur kernel is used to downsample images and synthesize training samples, which can be used to train a specific model for a given kernel and LR image. In~\cite{srmd}, the kernel and LR image are directly concatenated at the first layer of a DNN. Thus, the SR result can be closely correlated to both LR image and blur kernel. In~\cite{dpsr}, Zhang \etal propose a method based on the ADMM algorithm. They interpret this problem as MAP optimization and solve the data term and prior term alternately. A similar idea is adopted in~\cite{usr}. These methods can achieve remarkable performance as long as the ground-truth blur kernel is known. However, in real applications, the blur kernels are predicted by kernel-estimating methods, which are biased from the ground-truth ones. As we will illustrate in Sec~\ref{compare}, this bias will cause a serve performance drop when the two steps are combined together.
	
	\subsection{End-to-End Blind Super Resolution}
	End-to-end methods for blind SR are rarely studied before. In~\cite{ikc}, a kernel-estimation module and a non-blind-SR module are firstly integrated into a single blind SR method. It further proposes a correction module, which uses the super-resolved SR image to iteratively correct the estimated kernel. However,  the three modules in~\cite{ikc} are still trained in two steps, which is complicated and may restrict its performance. In the proposed method of our paper,  the kernel-estimation and SR modules are end-to-end optimized, which is not only much simpler but also can help the two modules get more compatible with each other and achieve better performance.
	
	\begin{figure*}
		\centering
		\includegraphics[width=\linewidth]{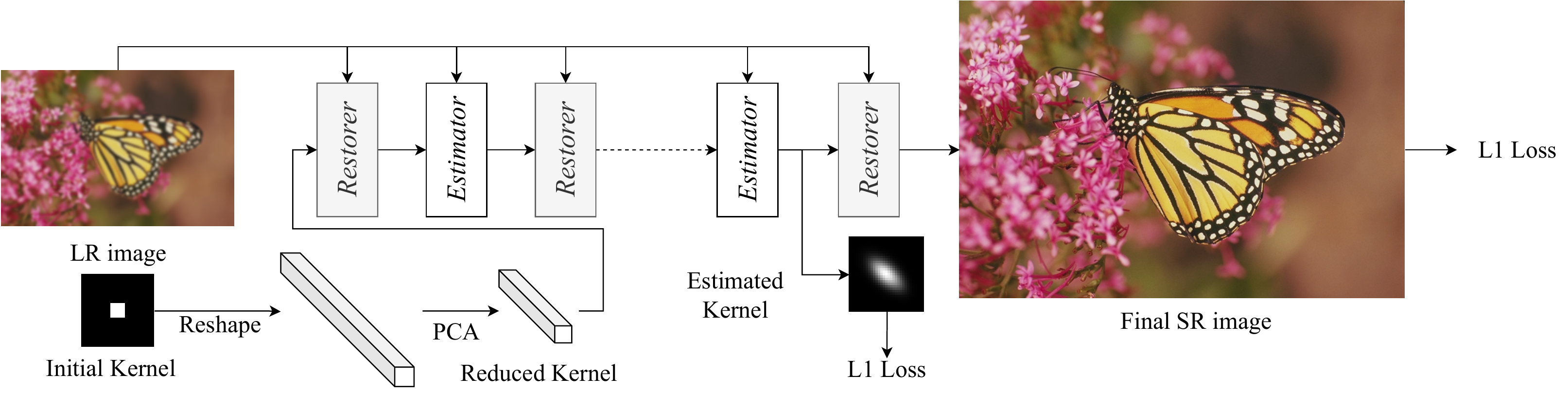}
		\caption{The overview of the deep alternating network (DAN).} \label{overview}
	\end{figure*}
	
	\section{End-to-End Alternating Optimization}
	In this section, we first illustrate the overall algorithm of our proposed and then go into the details. We start from the formulation of blind SR,  which helps us explain our method mathematically. The design details will be described at last. 
	\subsection{Formulation}\label{formulation}
	As shown in Equation~\ref{downsample}, there are three variables, \ie $\mathbf{x}$, $\mathbf{k}$ and $\mathbf{n}$, to be determined in blind SR problem.  From Equation~\ref{downsample} we can get
	\begin{equation}
		\mathbf{n} =  \mathbf{y} - (\mathbf{x}\otimes \mathbf{k})\downarrow_{s}.
	\end{equation}
	As $\mathbf{n}$ is assumed to be Gaussian noise with zero mean, the blind SR problem can be mathematically expressed an optimization problem in the Maximum A Posteriori (MAP) framework~\cite{zhang2017learning}: 
	\begin{equation}
		\mathop{\arg\min}_{\mathbf{k},\mathbf{x}}  \|\mathbf{y} - (\mathbf{x}\otimes \mathbf{k})\downarrow_{s} \|_1.
	\end{equation}
	Thus, the number of variables that need to be determined becomes $2$. However, this optimization problem is still  ill-posed and has an infinite number of solutions~\cite{baker2002limits}. To get it properly solved, some prior terms are usually added~\cite{selfdeblur,deblurpair}:
	\begin{equation}
		\mathop{\arg\min}_{\mathbf{k},\mathbf{x}}  \|\mathbf{y} - (\mathbf{x}\otimes \mathbf{k})\downarrow_{s} \|_1+ \phi(\mathbf{x}) + \psi(\mathbf{k}), 
	\end{equation}
	where $\phi(\mathbf{x})$ denotes the prior term for HR image, and $\psi(\mathbf{k})$ represents the prior term for blur kernel. In~\cite{BayesianSR}, Tipping \etal model the process of imaging and parameterize it with several unknown variables. They further assume that these unknown variables are subjected to high-dimensional Gaussian distributions. With the elaborated imaging model and strong assumptions,  they succeed to  solve this optimization problem directly. However, the imaging model or assumptions about unknown variables may be easily violated in real applications. On the other hand, without these strong assumptions, it is extremely difficult to solve this problem directly. 
	
	\subsection{Two-Step Solution}
	Given that the overall blind SR is difficult to be tackled, previous methods usually decompose this problem into two sequential steps:
	\begin{equation}
		\left \{
		\begin{aligned}
			\mathbf{k} &= M(\mathbf{y})\\
			\mathbf{x} &= \mathop{\arg\min}_{\mathbf{x}}  \|\mathbf{y} - (\mathbf{x}\otimes \mathbf{k})\downarrow_{s} \|_1 + \phi(\mathbf{x})
		\end{aligned}
		\right.
	\end{equation}
	where $M(\cdot)$ denotes the function that estimates $\mathbf{k}$ from $\mathbf{y}$, and the second step is usually solved by a non-blind SR method described in Sec~\ref{non-blind}. As we have mentioned in Sec~\ref{non-blind}, the two steps are independent research fields in most cases. Both of them only consider the performance under their own given conditions, while ignoring the overall performance. This two-step solution has its drawbacks in threefold. Firstly, this algorithm usually requires training of two or even more models, which is rather complicated. Secondly, $M(\cdot)$ can only utilize information from $\mathbf{y}$. However, this also an ill-posed problem:  $\mathbf{k}$ could not be properly solved without information from $\mathbf{x}$. At last, the non-blind SR model for the second step is trained with ground-truth kernels. While during testing, it can only have access to kernels estimated in the first step. The difference between ground-truth and estimated kernels will usually cause serve performance drop of the non-blind SR model~\cite{ikc}. 
	
	\subsection{Unfolding the Alternating Optimization}
	Towards the drawbacks of two-step solution, we propose an end-to-end network that can largely alleviate these issues. We still split it into two subproblems. However, instead of solving them sequentially, we adopt an alternating optimization algorithm, which restores the SR image and estimates the corresponding blur kernel alternately. The mathematical expression is
	\begin{equation}
		\left \{
		\begin{aligned}
			\mathbf{k}_{i+1} &=  \mathop{\arg\min}_{\mathbf{k}}  \|\mathbf{y} - (\mathbf{x}_{i}\otimes \mathbf{k})\downarrow_{s} \|_1 + \psi(\mathbf{k}) \\
			\mathbf{x}_{i+1} &= \mathop{\arg\min}_{\mathbf{x}}  \|\mathbf{y} - (\mathbf{x}\otimes \mathbf{k}_{i})\downarrow_{s} \|_1 + \phi(\mathbf{x}).
		\end{aligned}
		\right.
	\end{equation}
	We define two solvers, namely \textit{Estimator} and \textit{Restorer} for the two subproblems respectively. For \textit{Estimator}, there even has an analytic solution~\cite{wang2018generalized}. However, in current work, we choose to implement both solvers with convolutional neural modules. We have three reasons: 1) It is difficult to determine the appropriate analytic forms of the two prior terms. While neural modules are good at learning such priors implicitly~\cite{dip,asim2020blind,hou2019learning}. 2) Both modules tackle intermediate results, \ie $\mathbf{x}_{i}$ and $\mathbf{k}_{i}$ respectively, instead of ground-truth ones. Methods based on ground-truth assumptions may fail in this case. We also experimentally find that a neural-network-based \textit{Estimator} is more robust than the analytic solution in our method. 3) Once the neural modules are trained, it is easy for them to perform inference.
	
	Thus, we alternately the two subproblems with two neural modules. As shown in Figure~\ref{overview}, we fix the number of iterations as $T$ and unfold the iterating process to form an end-to-end trainable network, which is called a deep alternating network (DAN). We initialize the kernel by Dirac function, \ie the center of the kernel is one and zeros otherwise.  Following~\cite{ikc,srmd}, the kernel is also reshaped and then reduced by principal component analysis (PCA)~\cite{pca}. We set $T=4$ in practice and both modules are supervised only at the last iteration by L1 loss. The whole network could be well trained without any restrictions on intermediate results because the parameters of both modules are shared between different iterations. 
	
	In DAN, \textit{Estimator} takes both LR and SR images as inputs, which makes the estimation of blur kernel $\mathbf{k}$ much easier. More importantly, \textit{Restorer} is trained with the kernel estimated by \textit{Estimator}, instead of the ground-truth kernel as previous methods do. Thus, \textit{Restorer} could be more tolerant to the estimation error of \textit{Estimator} during testing. Besides, compared with previous two-step solutions, the results of both modules in DAN could be substantially improved, and it is likely for DAN to get better final results. Especially, in the case where the scale factor $s=1$, DAN becomes a deblurring network.
	
	\begin{figure*}[t]
		\centering
		\includegraphics[width=\linewidth]{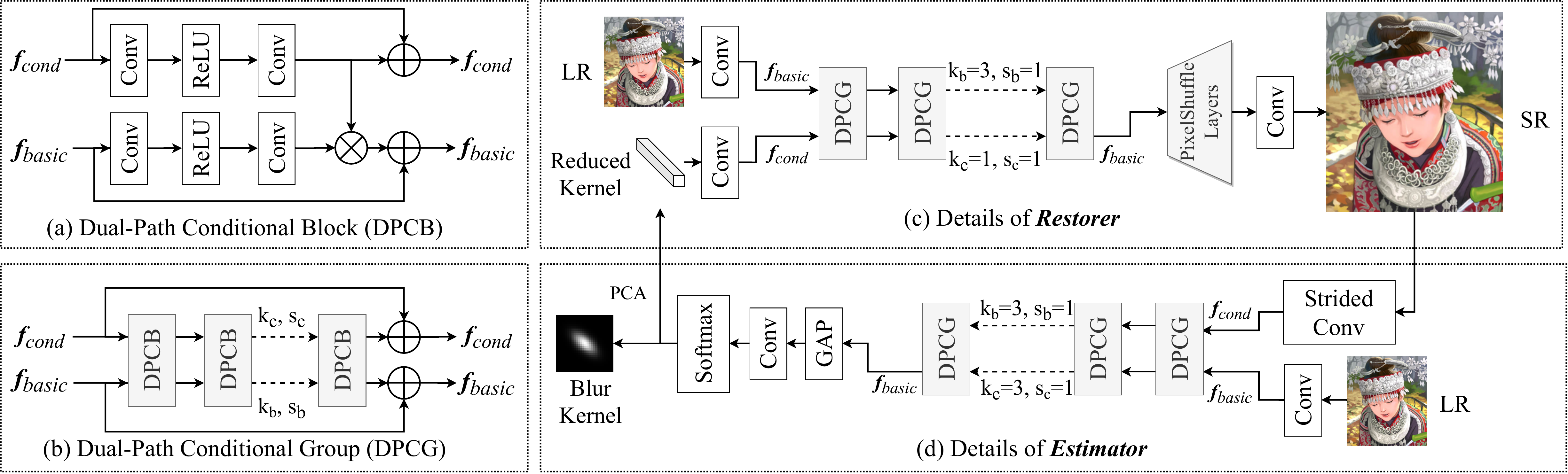}
		\caption{The details of  (a) dual-path conditional bock (DPCB), (b) dual-path conditional group (DPCG), (c) \textit{Restorer}, and (d) \textit{Estimator}. `GAP' denotes Global Average Pooling, $f_{basic}$ denote the basic input, and $f_{cond}$ denotes the conditional input.} \label{modules}
	\end{figure*}
	
	\subsection{Instantiate Convolutional Neural Modules}
	The most direct way to build the \textit{Estimator} and \textit{Restorer} is using the kernel estimation network and non-blind SR network in previous methods~\cite{ikc,usr}. However, on the one hand, those networks are too large to be directly combined together. On the other hand, the performance of our proposed method is more related to the compatibility between \textit{Estimator} and \textit{Restorer}. Architectures designed for cases where they are working alone may be not suitable for the case in DAN. Thus, in this section, we specially design the architectures for \textit{Estimator} and \textit{Restorer}.
	
	\subsubsection{Design of Basic Elements}~\label{basic_block}
	\noindent\textbf{Analysis.}
	Both modules in our network have two inputs. \textit{Estimator} takes LR and SR image, and \textit{Restorer} takes LR image and blur kernel as inputs. We define the LR image as the basic input, and the other one is the conditional input, \ie the blur kernel, and SR image is the conditional input of \textit{Restorer} and \textit{Estimator} respectively. During iterating, the basic inputs of both modules keep the same, but their conditional inputs are repeatedly updated. We claim that it is significantly important to keep the output of each module closely related to its conditional input. Otherwise, the iterating results will collapse to a fixed point at the first iteration. Specifically, if \textit{Estimator} outputs the same kernel regardless of the value of SR image, or \textit{Restorer} outputs the same SR image regardless of the value of blur kernel, their outputs will only depend on the basic input, and the results will keep the same during the iterating.
	
	\vspace{0.02\linewidth}
	\noindent\textbf{Conditional Residual Block.}
	In the conference version~\cite{dan},  a conditional residual block (CRB) is used to ensure the outputs of \textit{Estimator} and \textit{Restorer} are closely related to their conditional inputs. However, this block has three drawbacks: 1) In \textit{Restorer}, the conditional input, \ie he estimated kernels have to be expanded spatially to get concatenated with the LR features, which largely increases the computational cost. 2) Experiments show that the channel attention layer (CALayer) in CRB is time-consuming and will easily lead to gradient explosion, which slows down the inference and makes the training unstable. 3) All blocks in the network are conditioned by the same features, which may restrict the representing ability of the whole network.
	
	\vspace{0.02\linewidth}
	\noindent\textbf{Dual-Path Conditional Block.}
	To overcome the drawbacks of the conditional residual block,  we propose a dual-path conditional block (DPCB) in this paper. As shown in Figure~\ref{modules} (a),  there are two paths in DPCB, \ie conditional path (top one) and basic path (bottom one). we do not concatenate the conditional and basic paths directly.  Instead, they are independently processed firstly and then are multiplied to get correlated.  If  the conditional input has different spatial sizes as the basic input, it is expanded just before the multiplication. In this way, convolutions on the conditional input are performed before the spatial expansion, which saves much computation.  Besides,  we add skip connection on the conditional path, which enables the basic inputs at different depths are conditioned by different features. It may improve the representing ability of the whole module and enhance the final results. We also remove the channel attentional layer to accelerate the inference and stabilize the training.
	
	\vspace{0.02\linewidth}
	\noindent\textbf{Dua-Path Conditional Group.}
	We further adopt the residual in residual (RIR) structure proposed in~\cite{rcan}. As shown in~\ref{modules} (b), we add long skip connections when several DPCBs are sequentially stacked. These blocks form what we call dual-path conditional group (DPCG). These long skip connections could further help stabilize the training and enhance the results of very deep neural networks~\cite{rcan}. Since the conditional and basic paths are independently processed, the convolutional layers on the two paths can also have different configurations. As shown in Figure~\ref{modules} (b), we denote the kernel size and stride for the two paths as $k_c$, $k_b$ and $s_c$ and $s_b$ respectively.
	
	\subsubsection{\textit{Restorer}}
	The whole structure of \textit{Restorer} is shown in Figure~\ref{modules} (c). Both inputs are firstly mapped to have the same number of channels by a single convolutional layer respectively. The body of  \textit{Restorer} consists of only DPCGs.  The spatial size of the conditional input, \ie the reduced kernel, is $1\times1$. In this case, the conditional input needs to be expanded spatially to get multiplied with the basic input in the DPCB. Fortunately, the conditional input can maintain the $1\times1$ spatial size through the conditional path, which saves many computations than the conference verison~\cite{dan}. We use PixelShuffle~\cite{pixel_shuffle} layers to upscale the features to the desired size. In practice, \textit{Restorer} consists of $5$ DPCGs and each DPCG contains $10$ DPCBs. The number of channels in the body is set as $64$.
	
	\subsubsection{\textit{Estimator}}~\label{estimator}
	The whole structure of \textit{Estimator} is shown in Figure~\ref{modules} (d). The SR image super-resolved by \textit{Restorer} is firstly downscaled by a convolutional layer with stride $s$. Then the feature maps are used as the conditional input of \textit{Estimator}. The body of  \textit{Estimator} also consists of only DPCGs. The kernel sizes for both basic and conditional paths are set as $3\times3$. In practice, the body of \textit{Estimator} consists one DPCG, which contains $5$ DPCBs. The number of channels in the body is set as $32$.
	
	In the conference version~\cite{dan}, the \textit{Estimator} only predicts kernels in the reduced space and it is only supervised by the reduced kernel. There are two drawbacks to this design: 1) \textit{Estimator} can not predict complete kernels, \ie kernels before being transformed by PCA. Even the final SR result is good enough, we do not know how the blur kernel looks like. 2) Although the reduced kernel is well-supervised,  the complete kernel is not well-constrained. While according to~\cite{levin},  it is better to restrict the complete kernel to sum to one. which is important to the convergence of the whole algorithm~\cite{levin2006blind}. Thus, in current version, the \textit{Estimator} directly predicts all elements of the blur kernel, \ie the complete kernel. We further add a Softmax~\cite{softmax} layer at the end of \textit{Estimator}, which explicitly forces the complete kernel to sum to one.  Experiments in Sec~\ref{kernel_study} indicates that the predicted kernels of modified \textit{Estimator} have fewer visual distinctions with ground truth and smaller quantitative error.

	\begin{table*}[t]
		\centering
		\caption{Quantitative comparison with SOTA SR methods with Setting 1. The best two results are indicated in bold and underlined respectively.} \label{setting1}
		\resizebox{\linewidth}{!}{
			\begin{tabular}{cccccccccccc} 
				\hline
				\multirow{2}{*}{Method} & \multirow{2}{*}{Scale} & \multicolumn{2}{c}{Set5} & \multicolumn{2}{c}{Set14} & \multicolumn{2}{c}{BSD100} & \multicolumn{2}{c}{Urban100} & \multicolumn{2}{c}{Manga109} \\
				&                        & ~~PSNR~~       & ~~SSIM~~        & ~~PSNR~~        & ~~SSIM~~        & ~~PSNR~~        & ~~SSIM~~         & ~~PSNR~~         & ~~SSIM~~          & ~~PSNR~~         & ~~SSIM~~          \\ 
				\hline
				\hline
				Bicubic                 & \multirow{7}{*}{2}     
				& 28.82      & 0.8577      & 26.02       & 0.7634      & 25.92       & 0.7310       & 23.14        & 0.7258        & 25.60        & 0.8498        \\
				CARN~\cite{carn}                    &                        
				& 30.99      & 0.8779      & 28.10       & 0.7879      & 26.78       & 0.7286       & 25.27        & 0.7630        & 26.86        & 0.8606        \\
				Bicubic+ZSSR~\cite{zssr}                    &                        
				& 31.08      & 0.8786      & 28.35       & 0.7933      & 27.92       & 0.7632       & 25.25        & 0.7618        & 28.05        & 0.8769        \\
				\cite{pan}+CARN~\cite{carn}                 &                        
				& 24.20      & 0.7496      & 21.12       & 0.6170      & 22.69       & 0.6471       & 18.89        & 0.5895        & 21.54        & 0.7946        \\
				CARN~\cite{carn}+\cite{pan}                  &                        
				& 31.27      & 0.8974      & 29.03       & 0.8267      & 28.72       & 0.8033       & 25.62        & 0.7981        & 29.58        & 0.9134        \\
				IKC~\cite{ikc}                     &                        
				&37.19      &0.9526       &32.94       &0.9024        &31.51         &0.8790       &29.85          &0.8928       &36.93           &0.9667        \\
				DANv1 ~\cite{dan}                              &                   
				&\underline{37.34}         &\underline{0.9526}            &\underline{33.08}         &\underline{0.9041}       &\underline{31.76}         &\underline{0.8858}      &\underline{30.60}        &\underline{0.9060}         &\underline{37.23}             &\underline{0.9710}             \\ 
				DANv2                             &                   
				&\textbf{37.60}         &\textbf{0.9544}            &\textbf{33.44}         &\textbf{0.9094}            &\textbf{32.00}          &\textbf{0.8904}       &\textbf{31.43}              &\textbf{0.9174}       &\textbf{38.07}            &\textbf{0.9734}             \\ 
				\hline
				Bicubic                 & \multirow{7}{*}{3}     
				& 26.21      & 0.7766      & 24.01       & 0.6662      & 24.25       & 0.6356       & 21.39        & 0.6203        & 22.98        & 0.7576        \\
				CARN~\cite{carn}                    &                       
				& 27.26      & 0.7855      & 25.06       & 0.6676      & 25.85       & 0.6566       & 22.67        & 0.6323        & 23.84        & 0.7620        \\
				Bicubic+ZSSR~\cite{zssr}                    &                        
				& 28.25      & 0.7989      & 26.11       & 0.6942      & 26.06       & 0.6633       & 23.26        & 0.6534        & 25.19        & 0.7914        \\
				\cite{pan}+CARN~\cite{carn}                  &                        
				& 19.05      & 0.5226      & 17.61       & 0.4558      & 20.51       & 0.5331       & 16.72        & 0.4578        & 18.38        & 0.6118        \\
				CARN~\cite{carn}+\cite{pan}                &                        
				& 30.31      & 0.8562      & 2757        & 0.7531      & 27.14       & 0.7152       & 24.45        & 0.7241        & 27.67        & 0.8592        \\
				IKC~\cite{ikc}                     &                       
				&33.06       &0.9146       &29.38       &0.8233       &28.53        &0.7899        &24.43         &0.8302         &32.43         &0.9316        \\
				DANv1~\cite{dan}                               &                        
				&\underline{34.04}            &\underline{0.9199}          &\underline{30.09}             &\underline{0.8287}           &\underline{28.94}           &\underline{0.7919}              &\underline{27.65}              &\underline{0.8352}              &\underline{33.16}           &\underline{0.9382}               \\ 
				DANv2                            &                        
				&\textbf{34.19}            &\textbf{0.9209}          &\textbf{30.20}             &\textbf{0.8309}           &\textbf{29.03}            &\textbf{0.7948}              &\textbf{27.83}         &\textbf{0.8395}              &\textbf{33.28}              &\textbf{0.9400}               \\ 
				\hline
				Bicubic                 & \multirow{7}{*}{4}     
				& 24.57      & 0.7108      & 22.79       & 0.6032      & 23.29       & 0.5786       & 20.35        & 0.5532        & 21.50        & 0.6933        \\
				CARN~\cite{carn}                    &                        
				& 26.57      & 0.7420      & 24.62       & 0.6226      & 24.79       & 0.5963       & 22.17        & 0.5865        & 21.85        & 0.6834        \\
				Bicubic+ZSSR~\cite{zssr}                    &                       
				& 26.45      & 0.7279      & 24.78       & 0.6268      & 24.97       & 0.5989       & 22.11        & 0.5805        & 23.53        & 0.7240        \\
				\cite{pan}+CARN~\cite{carn}                  &                       
				& 18.10      & 0.4843      & 16.59       & 0.3994      & 18.46       & 0.4481       & 15.47        & 0.3872        & 16.78        & 0.5371        \\
				CARN~\cite{carn}+\cite{pan}                 &                        
				& 28.69      & 0.8092      & 26.40       & 0.6926      & 26.10       & 0.6528       & 23.46        & 0.6597        & 25.84        & 0.8035        \\
				IKC~\cite{ikc}                    &                        
				&31.67       &0.8829      &28.31         &0.7643       & 27.37       &0.7192       &25.33          &0.7504          &28.91           &0.8782        \\
				DANv1~\cite{dan}              &                        
				&\underline{31.89}       &\underline{0.8864}      &\underline{28.42}       &\underline{0.7687}      &\underline{27.51}        &\underline{0.7248}      &\underline{25.86}     &\underline{0.7721} &\underline{30.50} &\underline{0.9037}             \\ 
				DANv2              &                        
				&\textbf{32.00}       &\textbf{0.8885}     &\textbf{28.50}       &\textbf{0.7715}     & \textbf{27.56}        &\textbf{0.7277}      &\textbf{25.94}     &\textbf{0.7748}  &\textbf{30.45} &\textbf{0.9037}             \\ 
				\hline
		\end{tabular}}
	\end{table*}
	\begin{figure*}[t]
		\centering
		\includegraphics[width=\linewidth]{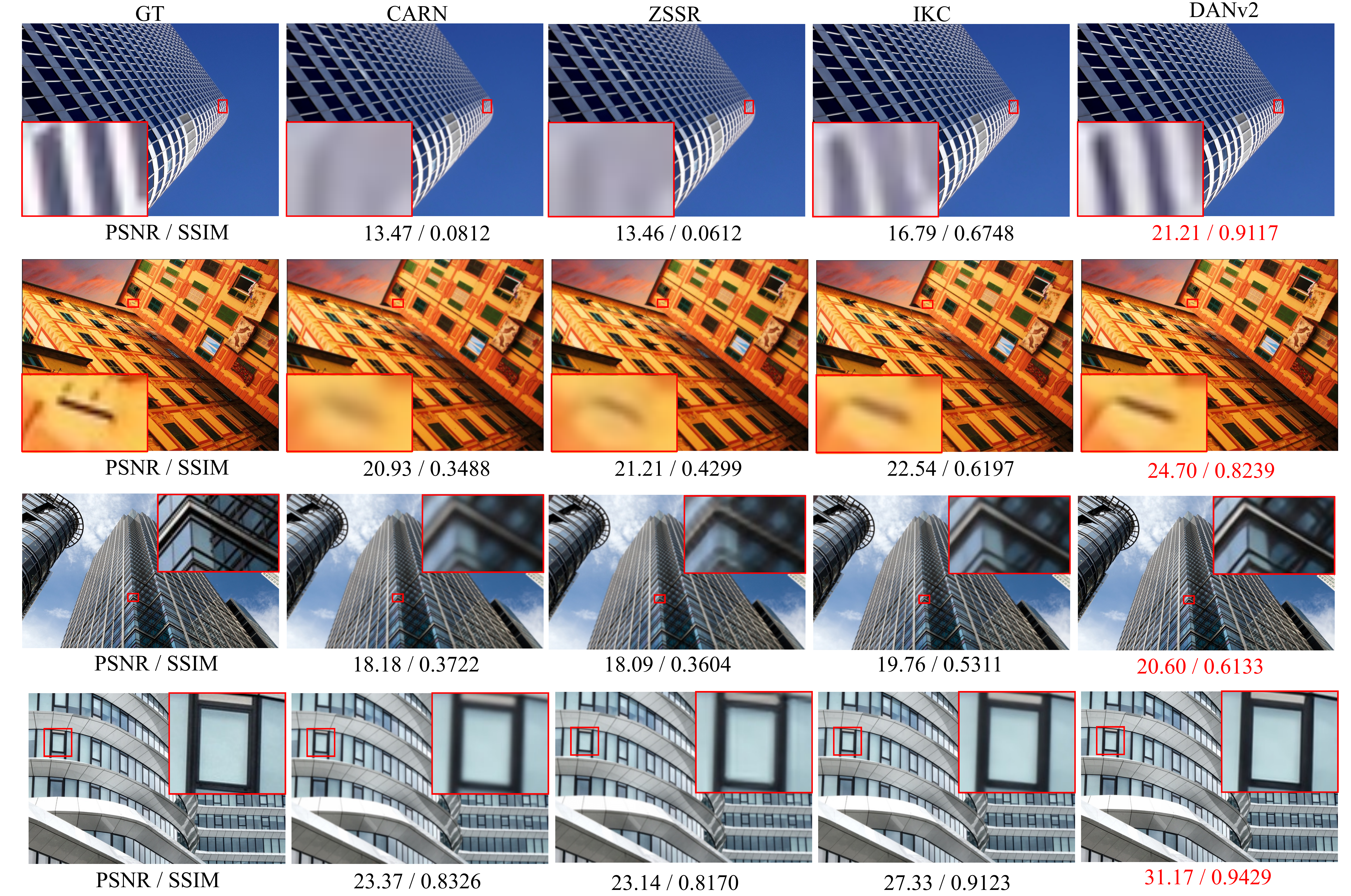}
		\caption{Visual results of  \textit{img 005}, \textit{img 013}, \textit{img 047} and \textit{img 052} in Urban100. The $\sigma$ of blur kernel is $1.8$. Best viewed in color.}\label{visual_setting1}
	\end{figure*}
	\begin{figure}[t]
		\centering
		\includegraphics[width=\linewidth]{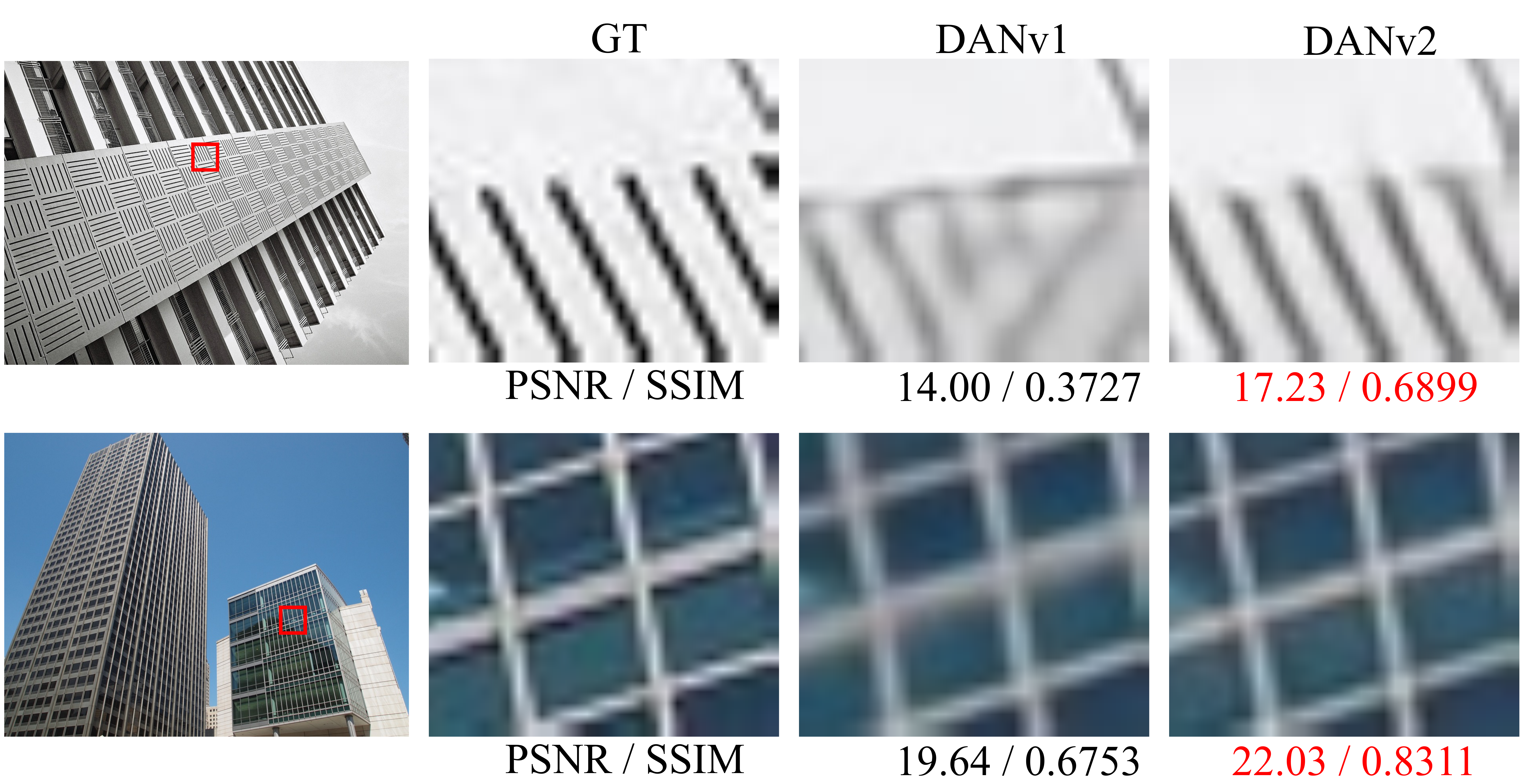}
		\caption{Visual results of \textit{img 092} and \textit{img 096} in Urban100. The $\sigma$ of blur kernel is $1.8$}\label{setting1_dan_comp}
	\end{figure}
	
	\section {Experiments}
	To fully investigate the proposed method, experiments are performed on both synthetic and real images. In experiments on synthetic images, we evaluate its quantitative results under different settings and perform controlled experiments to help analyze the proposed method. In experiments on real images, we provide a qualitative comparison to demonstrate the effectiveness of the proposed method. 
	
	\subsection{Experiments on Synthetic Images}
	
	To fully investigate the proposed method,  extensive experiments are performed under two different degradation settings. Setting 1 only focuses on cases of isotropic Gaussian blur kernels. In this case, different blur kernels can be quantitatively compared, which can help study the influence of blur kernels. Setting 2 focuses on cases of more general and irregular blur kernels.  Intuitively,  Setting 2 is relatively more difficult and can help study the performance of the proposed method.
	
	\vspace{0.02\linewidth}\noindent
	\textbf{Setting 1.} Following the setting in~\cite{ikc}, the kernel size is set as $21$. During training,  the kernel width is uniformly sampled in [0.2, 4.0], [0.2, 3.0] and [0.2, 2.0] for scale factors $4$, $3$ and $2$ respectively. For quantitative evaluation, we collect HR images from the commonly used benchmark datasets, \ie Set5~\cite{set5}, Set14~\cite{set14}, Urban100~\cite{urban100}, BSD100~\cite{bsd100} and Manga109~\cite{manga109}. Since determined kernels are needed for reasonable comparison, we uniformly choose 8 kernels, denoted as \textit{Gaussian8}, from range [1.8, 3.2], [1.35, 2.40] and [0.80, 1.60] for scale factors $\times4$, $\times3$ and $\times2$ respectively. The HR images are first blurred by the selected blur kernels and then downsampled to form synthetic test images. 
	
	\begin{table*}[t]
		\centering
		\caption{Quantitative comparison with SOTA SR methods with Setting 2. The best two results are indicated in bold and underlined respectively.} \label{setting2}
		\footnotesize 
		\begin{tabular}{cccccc}
			\hline
			\multirow{3}{*}{Types}   
			& \multirow{3}{*}{Method}     & \multicolumn{4}{c}{Scale}                     \\
			&                             & \multicolumn{2}{c}{2} & \multicolumn{2}{c}{4} \\
			&                             & ~~PSNR~~      & ~~SSIM~~      & ~~PSNR~~      & ~~SSIM~~      \\ 
			\hline
			\hline
			\multirow{4}{*}{Class 1}
			& Bicubic                     & 28.73     & 0.8040    & 25.33     & 0.6795    \\
			& Bicubic kernel + ZSSR~\cite{zssr}       & 29.10     & 0.8215    & 25.61     & 0.6911    \\
			& EDSR~\cite{edsr}                    & 29.17     & 0.8216    & 25.64     & 0.6928    \\
			& RCAN~\cite{rcan}                  & 29.20     & 0.8223    & 25.66     & 0.6936    \\ \hline
			\multirow{5}{*}{Class 2} 
			&~~PDN~\cite{2018ntire} - 1st in NTIRE'19 track4             & /   & /     & 26.34     & 0.7190    \\
			& ~~WDSR~\cite{wdsr} - 1st in NTIIRE'19 track2~~   & /   & /     & 21.55     & 0.6841    \\
			& WDSR~\cite{wdsr} - 1st in NTIRE'19 track3            & /   & /      & 21.54     & 0.7016    \\
			&~~WDSR~\cite{wdsr} - 2nd in NTIRE'19 track4~~    &  /   & /       & 25.64     & 0.7144    \\ 
			&Ji \etal ~\cite{ji2020real} - 1st in NITRE'20 track 1 &  /  & /     & 25.43     & 0.6907 \\ \hline
			\multirow{6}{*}{Class 3}
			& Cornillere \etal~\cite{siga}& 29.46     & 0.8474    & /     & /   \\
			& Michaeli \etal~\cite{nonpara} + SRMD ~\cite{srmd}              & 25.51     & 0.8083    & 23.34     & 0.6530    \\
			& Michaeli \etal~\cite{nonpara} + ZSSR~\cite{zssr}               & 29.37     & 0.8370    & 26.09     & 0.7138    \\
			& KernelGAN~\cite{kernel_gan} + SRMD~\cite{srmd}            & 29.57     & 0.8564    & 25.71     & 0.7265    \\
			& KernelGAN~\cite{kernel_gan} + USRNet~\cite{usr}             & /          &   /             & 20.06     & 0.5359 \\
			& KernelGAN ~\cite{kernel_gan}+ ZSSR~\cite{zssr}           &30.36     &0.8669    &26.81     &0.7316    \\ \hline
			\multirow{2}{*}{Ours} 
			& DANv1                      &\underline{32.56}&\underline{0.8997}          &\underline{27.55} &\underline{0.7582}        \\ 
			& DANv2                      &\textbf{32.58}&\textbf{0.9048}          &\textbf{28.74} &\textbf{0.7893}        \\ 
			\hline
		\end{tabular}
	\end{table*}
	\begin{figure*}[t]
		\centering
		\includegraphics[width=\linewidth]{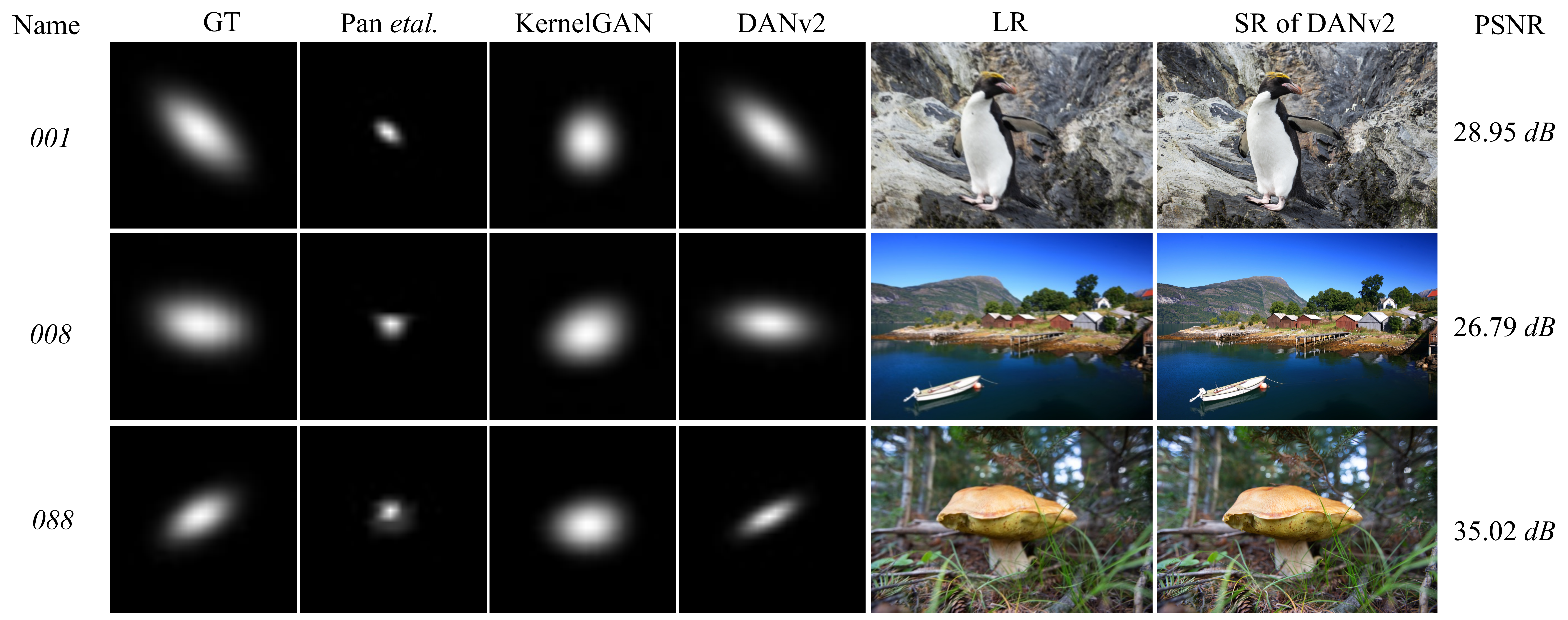}
		\caption{Visual results on DIV2KRK. From the most left to the most right column are respectively ground-truth kernels, kernels estimated by Pan \etal~\cite{pan}, kernels estimated by KernelGAN~\cite{kernel_gan}, kernels estimated by DANv2, corresponding LR images, and SR results restored by DANv2. From the top to the bottom row are respectively results of Image \textit{001}, \textit{008}, and \textit{088}. We also list out the PSNR results of SR image restored by DANv2. Best viewed in color.}\label{vis_kernel}
	\end{figure*}
	\begin{figure*}[t]
		\centering
		\includegraphics[width=\linewidth]{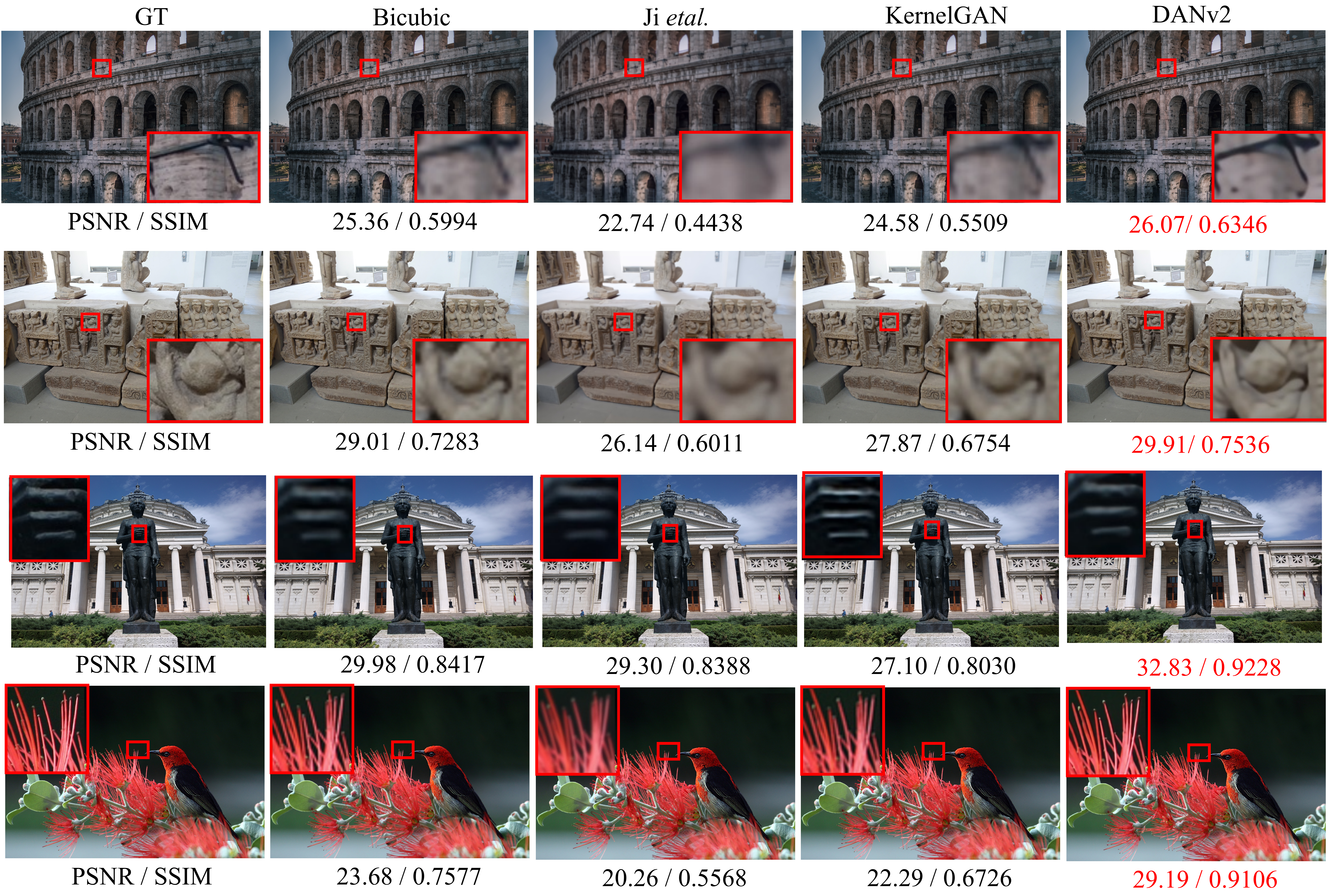}
		\centering{\caption{Visual results of  \textit{img 864}, \textit{img 816}, \textit{img 812} and \textit{img 853} in DIV2KRK. Best viewed in color.}\label{visual_setting2}
		} 
	\end{figure*}
	\begin{figure}[t]
		\centering
		\includegraphics[width=\linewidth]{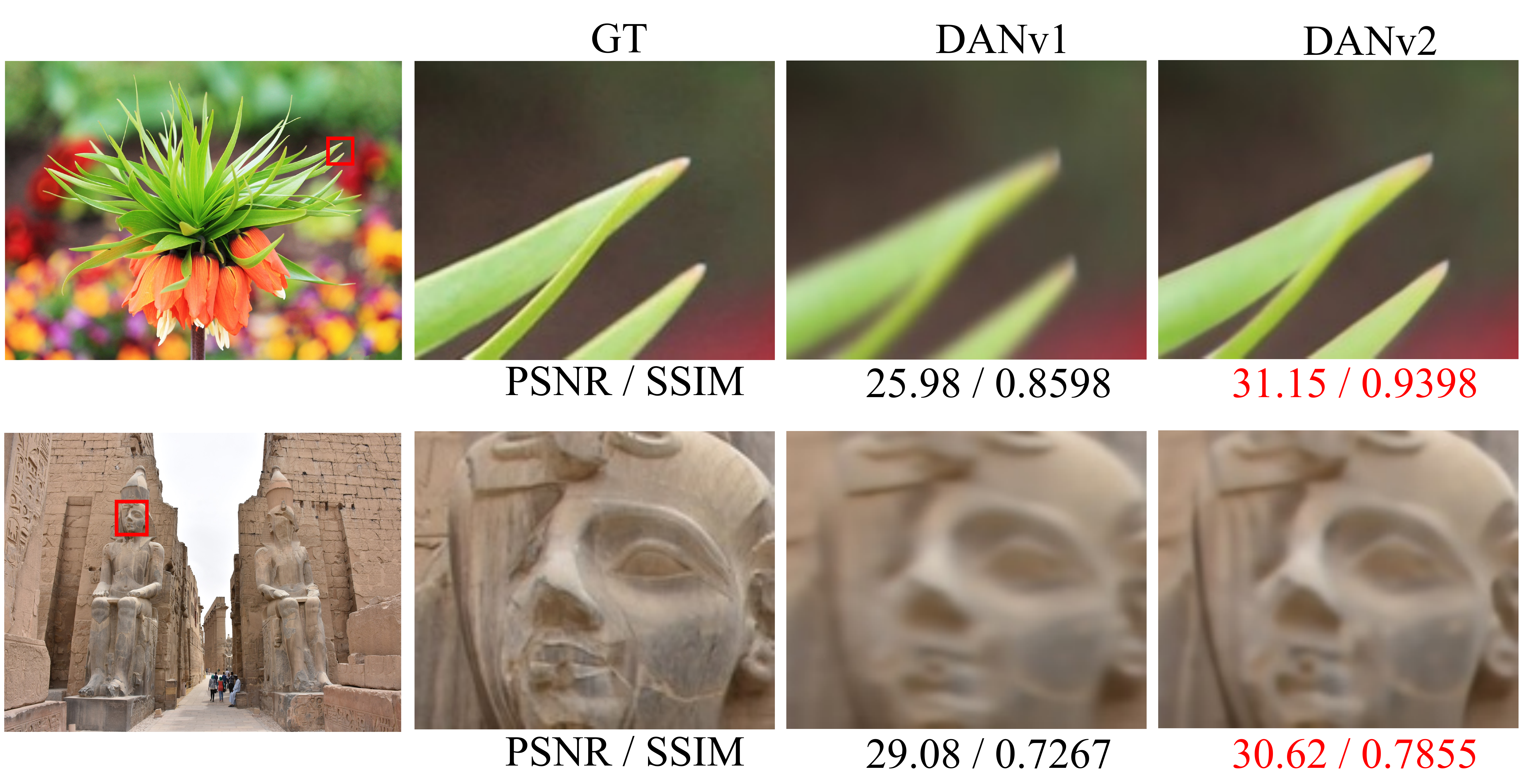}
		\caption{Visual results of \textit{img 003} and \textit{img 074} in Urban100. The width of blur kernel is $1.8$. Best viewed in color.}\label{setting2_dan_comp}
	\end{figure}
	\begin{figure}[t]
		\centering
		\includegraphics[width=\linewidth]{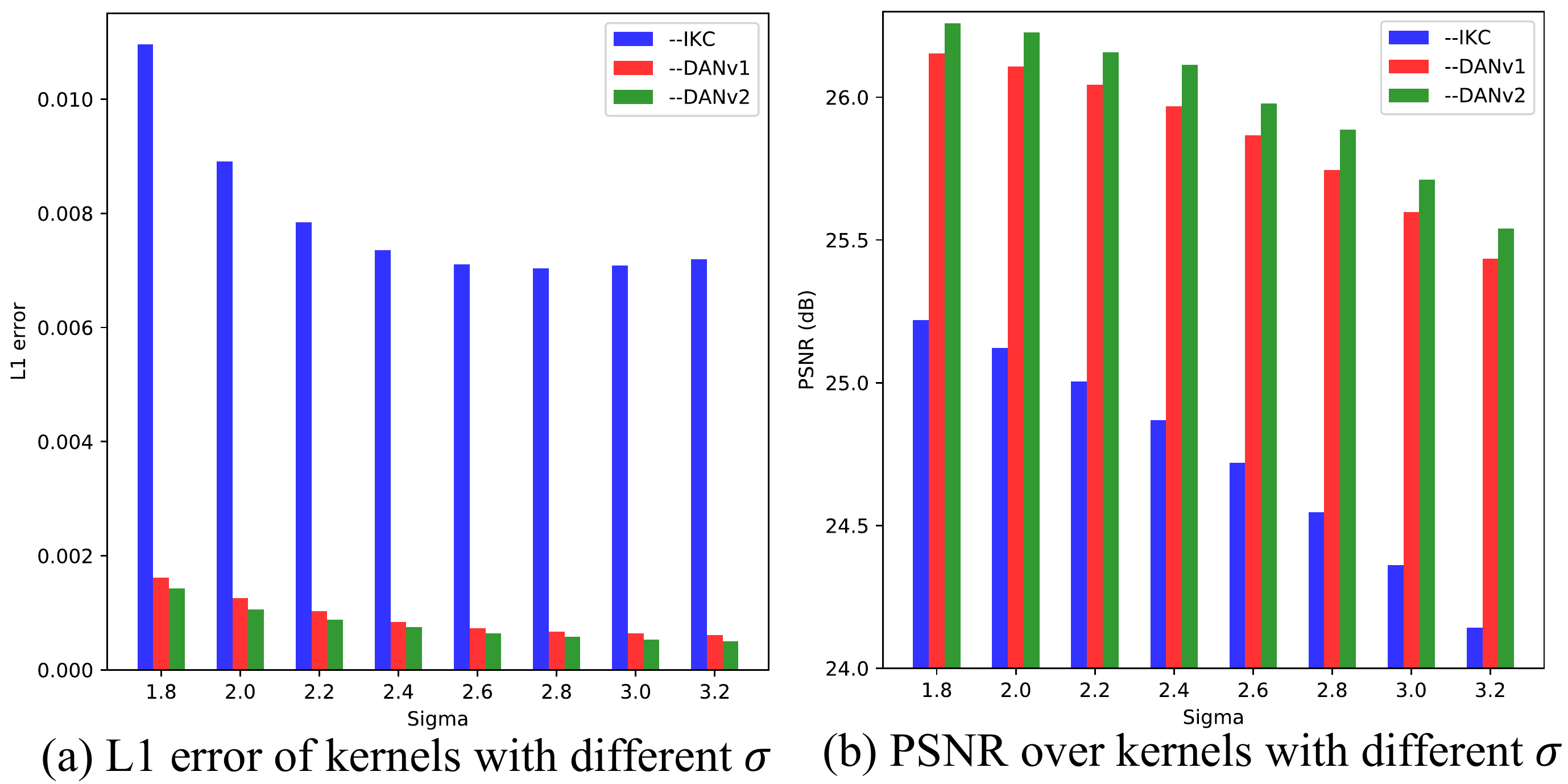}
		\caption{The L1 error of predicted kernels with different $\sigma$ (left) and PSNR results with respect to kernels with different $\sigma$ (right).}\label{kernel_results}
	\end{figure} 
	
	\vspace{0.02\linewidth}\noindent
	\textbf{Setting 2.} Following the setting in~\cite{kernel_gan},  we set the kernel sizes as $11$ and $31$ for scale $\times2$ and $\times4$ respectively. We firstly generate anisotropic Gaussian kernels. The lengths of both axes are uniformly distributed in $(0.6, 5)$, rotated by a random angle uniformly distributed in [$-\pi$, $\pi$]. To deviate from a regular Gaussian, we further apply uniform multiplicative noise (up to 25\% of each pixel value of the kernel) and normalize it to sum to one. For testing, we use the benchmark dataset DIV2KRK that is used in~\cite{kernel_gan}.
	
	\vspace{0.02\linewidth}\noindent
	\textbf{Data.} For both settings, we collect $3450$ HR images from DIV2K~\cite{div2k} and Flickr2K~\cite{flickr2k} as training set. We firstly crop all HR images to patches of $256\times256$ and use them to synthesize training pairs on the fly. The synthesized pairs are then further cropped such that the sizes of LR images are $64\times64$ for all scale factors. 
	
	\vspace{0.02\linewidth}\noindent
	\textbf{Training.} The batch sizes for all models are $64$. All models are trained for $4\times10^{5}$ iterations. We use Adam~\cite{adam} as our optimizer, with $\beta_1=0.9$, $\beta_2=0.99$. The initial learning rate is $4\times10^{-4}$, and will decay by half at every $2\times10^5$ iterations. All models are trained on $8$ RTX2080Ti GPUs.
	
	\vspace{0.02\linewidth}\noindent
	\textbf{Evaluation metric.} All methods are evaluated by PSNR and SSIM~\cite{ssim}. Both metrics are calculated on the Y channel (\ie luminance) of transformed YCbCr space.
	
	\subsubsection{Quantitative Comparisons}~\label{compare}
	In this section, we provide quantitative results of different methods under different settings.
	
	\vspace{0.02\linewidth}\noindent
	\textbf{Setting 1.}
	For the first setting, we evaluate our method on test images synthesized by \textit{Gaussian8} kernels. We denote DAN in the conference version~\cite{dan} as DANv1 and the DAN in current paper as DANv2. We mainly compare our results with ZSSR~\cite{zssr} (using bicubic kernel)  and IKC~\cite{ikc}. We also include a comparison with CARN~\cite{carn}. Since it is not designed for blind SR, we perform the deblurring method~\cite{pan} before or after CARN. The results in Table~\ref{setting1}. 
	
	Despite that CARN achieves remarkable results in the context of bicubic downsampling, it suffers severe performance drop when applied to images with unknown blur kernels. Its performance is largely improved when it is followed by a deblurring method,  but still inferior to that of blind-SR methods. ZSSR trains a specific network for each single tested image by utilizing the internal patch recurrence. However, ZSSR has an in-born drawback: the training samples for each image are limited, and thus it cannot learn a good prior for HR images. IKC is also a two-step solution for blind SR. Although the accuracy of the estimated kernel is largely improved in IKC, the final result is still suboptimal.
	
	Both DANv1 and DANv2 are trained in an end-to-end manner, which is not only much easier to be trained than two-step solutions but also more likely to reach a better optimum point. As shown in Table~\ref{setting1}, they outperform other methods by a large margin. Specially, DANv1 outperforms IKC by $3.22dB$ on Urban100 for scale $\times3$. This comparison indicates the importance of end-to-end training in blind SR. On the other hand, DANv2 is also improved a lot on the basis of DANv1. It suggests that the optimized structures of \textit{Restorer} and \textit{Estimator} are better than the conference version.
	
	\vspace{0.02\linewidth}\noindent
	\textbf{Setting 2.}
	The second setting involves irregular blur kernels, which are more general, but also more difficult to solve. For Setting 2, we mainly compare methods of three different classes: \textit{i}) SOTA SR algorithms trained on bicubically downsampled images such as EDSR~\cite{edsr} and RCAN~\cite{rcan} , \textit{ii}) blind SR methods designed for NTIRE competition such as PDN~\cite{2018ntire} and WDSR~\cite{wdsr}, \textit{iii}) the two-step solutions, \ie the combination of a kernel estimation method and a non-blind SR method, such as Kernel-GAN~\cite{kernel_gan} and ZSSR~\cite{zssr}. The PSNR and SSIM results on the Y channel are shown in Table~\ref{setting2}. 
	
	Similarly, the performance of methods trained on bicubically downsampled images is limited by the domain gap. Thus, their results are only slightly better than that of interpolation. The methods in Class 2 are trained on synthesized images provided in the NTIRE competition. Although these methods achieve remarkable results in the competition, they still cannot generalize well to irregular blur kernels. 
	
	The comparison between methods of Class 3 can enlighten us a lot. Specifically, USRNet~\cite{usr} achieves remarkable results when GT kernels are provided, and KernelGAN also performs well on kernel estimation. However, when they are combined together, as shown in Table~\ref{setting2}, the final SR results are worse than most other methods. This indicates that it is important for the \textit{Estimator} and \textit{Restorer} to be compatible with each other. Additionally, although a better kernel-estimation method can benefit the SR results, the overall performance is still largely inferior to that of both DANv1 and DANv2. This comparison also indicates the importance of end-to-end training for blind SR. Compared with DANv1,  the performance of DANv2 is further improved. Specially, DANv2 outperforms DANv1 by $1.19 dB$ for scale $\times4$. On the one hand, DPCB largely improves the representing ability of  DANv2. On the other hand, DANv2 can be trained more stably than DANv1. Thus it can be better optimized and achieve better results. 
	
	\subsubsection{Qualitative Comparisons with Other Methods}
	In this section, we provide some visual results of different methods under different settings for qualitative comparisons.
	
	\vspace{0.02\linewidth}\noindent
	\textbf{Setting 1.}
	The visual results of \textit{img 005}, \textit{img 013}, \textit{img 047} and \textit{img 052} in Urban100 are shown in Figure~\ref{visual_setting1} for comparisons between DAN and other methods. As one can see,  ZSSR and CARN even cannot restore clear edges. IKC performs better, but the edges are severely blurred. DANv2 restores sharper edges and simultaneously alleviates the blurriness. This comparison indicates that DAN could produce more visually pleasant SR images. For the qualitative comparisons between DANv1 and DANv2, we need to focus on harder cases. Because for relatively easier cases, both models perform well enough and their results are hard to be visually distinguished. We provide their results of \textit{img 092} and \textit{img 096} in Urban100 for comparisons. As shown in Figure~\ref{setting1_dan_comp}, it is likely for DANv1 to mix the stripes of different directions during the super-resolving processing. While DANv2 may be more stable for such cases.
	
	\vspace{0.02\linewidth}\noindent
	\textbf{Setting 2.}
	The visual results of \textit{img 864}, \textit{img 816}, \textit{img 812} and \textit{img 853} in Urban100 are shown in Figure~\ref{visual_setting2} for comparisons between DAN and other methods. We need to note that Bicubic interpolation is actually a strong baseline in blind SR. Although KernelGAN +ZSSR and Ji \etal can have better overall results on DIV2KRK, Bicubic interpolation can still outperform them in many cases. As indicated in the figure, compared with the other three methods, the SR images produced by DAN are much sharper and cleaner. We also provide individual comparisons between DANv1 and DANv2 in Figure~\ref{setting2_dan_comp}. As one can see, the SR images of DANv1 are still slightly blurred, while those of DANv2 are much cleaner.
	
	\subsubsection{Study of Estimated Kernels}~\label{kernel_study}
	\textbf{Accuracy.}
	We calculate the L1 error of predicted kernels to quantitatively evaluate their accuracy. As we want to investigate the performance over different kernels,  we choose to measure the predicted kernels in Setting 1, because different kernels in Setting 1 can be classified via their standard deviation $\sigma$. we calculate their L1 errors in the reduced space, and the results on Urban100 are shown in Figure~\ref{kernel_results} (a). As one can see that the L1 errors of reduced kernels predicted by DANv1 and DANv2 are much lower than that of IKC. It suggests that the overall improvements of DAN may partially come from more accurate predicted kernels. We need to note that DANv2 predicts more accurate kernels than DANv1, which demonstrates the modifications on \textit{Estimator} in Sec~\ref{estimator}. We also plot the PSNR results with respect to kernels with different $\sigma$ in Figure~\ref{kernel_results} (b). As $\sigma$ increases, the performance gap between IKC and DAN also becomes larger. It indicates that DAN may have better generalization ability.
	
	\vspace{0.02\linewidth}\noindent
	\textbf{Visualization.}
	Compared with DANv1, DANv2 directly predicts the complete blur kernel, instead of in the reduced space. It enables us to visualize the estimated kernels. In this section, we visualize some estimated kernels to qualitatively measure the performance of \textit{Estimator}. Since Gaussian kernels in Setting 1 are hard to be visually distinguished, we choose to visualize estimated kernels on DIV2KRK for scale factor $\times4$. The irregular kernels of DIV2KRK are more difficult to be estimated and the performances of different methods are easier to be visually measured. We use the results of  KernelGAN~\cite{kernel_gan} and Pan \etal~\cite{pan} as comparisons. As shown in Figure~\ref{vis_kernel},  kernels estimated by Pan \etal are collapsed to the central area. It indicates that this method fails in estimating relatively large kernels. The kernels estimated by KernelGAN are likely to be isotropic and look very different from the ground-truth kernels. Compared with these two methods, DAN can estimate the kernel much more accurately, even if the ground-truth kernels are highly anisotropic. 
	
	\subsubsection{Non-blind Setting}
	In this section, we replace the estimated kernel with ground truth (GT) to further investigate the influence of \textit{Estimator}. If GT kernels are provided, the iterating processing becomes meaningless. Thus we test the \textit{Restorer} with just once forward propagation. The tested results for Setting 1 are shown in Table~\ref{gt_kernel}. The result almost keeps unchanged and sometimes even gets worse when GT kernels are provided. It indicates that \textit{Predictor} may have already satisfied the requirements of \textit{Restorer}, and the superiority of DAN also partially comes from the good cooperation between its \textit{Predictor} and \textit{Restorer}.
	
	\begin{table}[h]
		\centering
		\setlength{\tabcolsep}{0.2cm}
		\caption{PSNR results when GT kernels are provided.}\label{gt_kernel}
		\resizebox{\linewidth}{!}{
			\begin{tabular}{c|cccccc}
				\hline
				Methods & Set5  & Set14 & B100  & Urban100 & Manga109 \\ \hline
				DANv2 & 32.00 & 28.50 & 27.56 & 25.94    & 30.45    \\
				DANv2(GT kernel) & 31.98 & 28.49 & 27.56 & 25.95    & 30.46    \\ \hline
			\end{tabular}
		}
	\end{table}
	
	\subsubsection{Ablation Study of Network Architectures}
	In this section, we investigate the influences of different architectures, including DPCB, DPCG, and Softmax layer in \textit{Estimator}. We use DAN of the conference version, \ie DANv1, as the baseline, which is denoted as experiment $A$. In experiment $B$, we replace the conditional residual block in DANv1 with DPCB. To control the model size,  the number of blocks is increased from $40$ to $50$. In experiment $C$, we further add long skip connections. We introduce the Softmax layer to \textit{Estimator} in experiment $D$, and the network finally becomes DANv2. We report the results of different experiments on Set14 in Setting 1. As shown in Table~\ref{arch_aba}, compared with the original conditional residual block in DANv1, DPCB and DPCG can improve the results by $0.05dB$.  The Softmax layer in \textit{Estimator} can further improve the results by $0.03dB$. It indicates that it helps to explicitly restrict the estimated kernels to sum to one.
	
	\begin{table}[t]
		\centering
		\caption{Ablation study results of network acchitectures. Results are reported as average PSNR on Urban100 in Setting 1.}\label{arch_aba}
		\setlength{\tabcolsep}{0.4cm}
		\resizebox{\linewidth}{!}{
			\begin{tabular}{c|ccc|c}
				\hline
				Exp.& DPCB & DPCG & Softmax & Results \\ \hline
				$A$ &      &      &         & $25.86$        \\
				$B $& $\surd$&      &         &$25.89$      \\
				$C $& $\surd$ &$\surd$ &         &$25.91$         \\
				$D $&$\surd$&$\surd$&$\surd$&$25.94$        \\ \hline
		\end{tabular}}
		
	\end{table}
	
	\begin{figure*}[t]
		\centering
		\includegraphics[width=\linewidth]{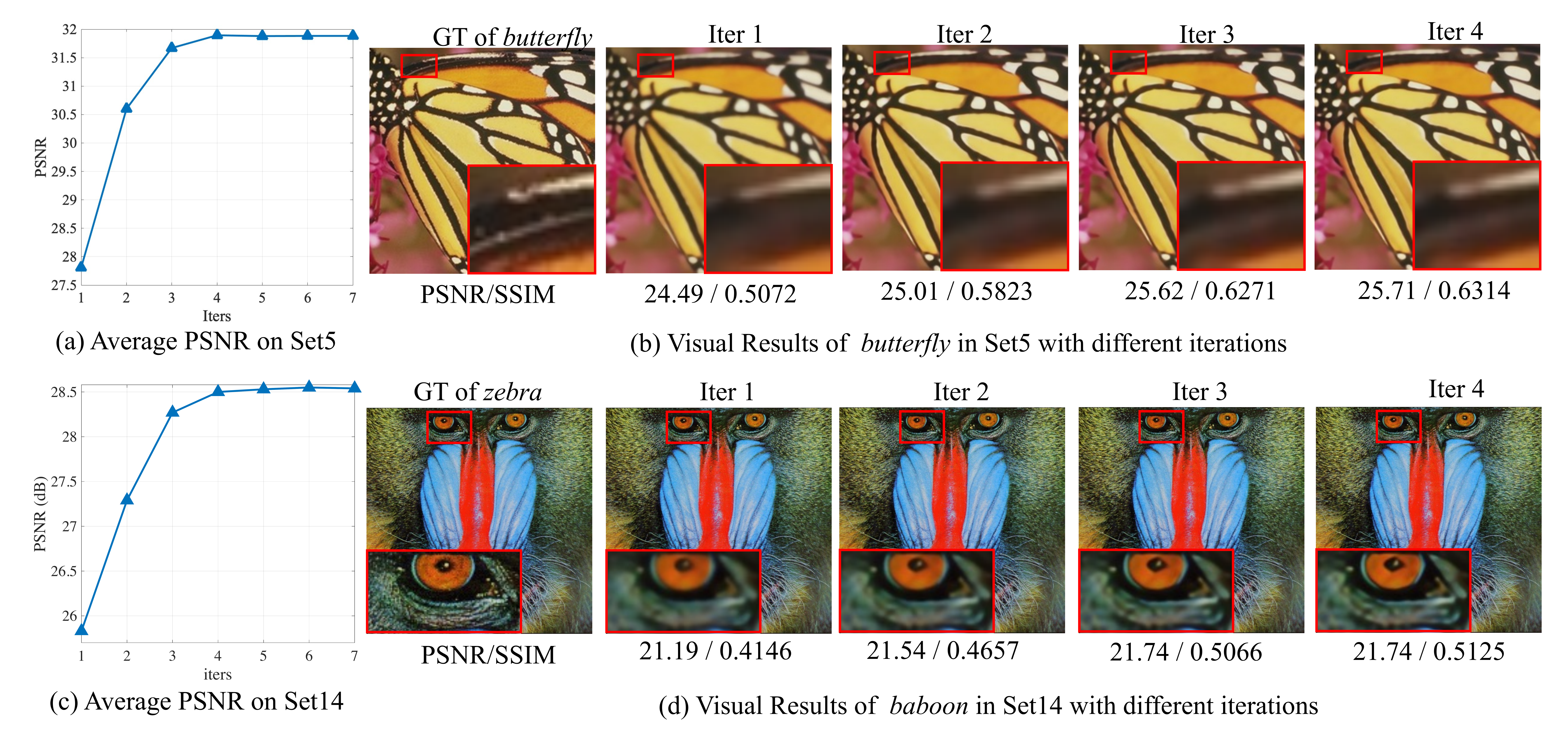}
		\centering
		\caption{PSNR and visual results with different iterations during testing on Set5 and Set14.}\label{iter}
	\end{figure*}
	\begin{figure*}[t]
		\centering
		\includegraphics[width=\linewidth]{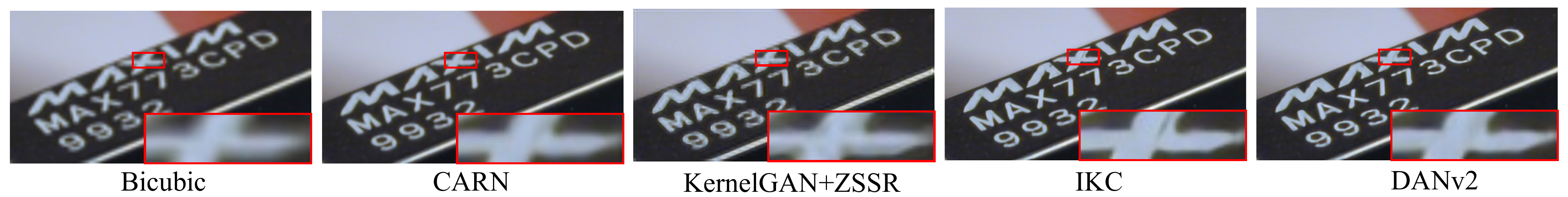}
		\caption{Visual results on real image \textit{chip}.}\label{chip}
	\end{figure*}
	
	\subsubsection{Study of Iterations}
	After the model is trained, we also change the number of iterations to see whether the two modules have learned the property of convergence or just have `remembered' the iteration number. The model is trained with $4$ iterations, but during testing, we increase the iteration number from $1$ to $7$. As shown in Figure~\ref{iter} (a) and (c), the average PSNR results on Set5 and Set14 firstly increase rapidly and then gradually converge. It should be noted that when we iterate more times than training, the performance does not become worse, and sometimes even becomes better. For example, the average PSNR on Set14 is $20.43dB$ when the iteration number is $5$, higher than $20.42dB$ when we iterate $4$ times. Although the incremental is relatively small, it suggests that the two modules may have learned to cooperate with each other, instead of solving this problem like ordinary end-to-end networks, in which cases, the performance will drop significantly when the setting of testing is different from that of training. It also suggests that the estimation error of intermediate results does not destroy the convergence of DAN. In other words, DAN is robust to various estimation errors.
	
	\begin{table}[h]
		\centering
		\caption{Comparisons on model complexities and inference speed of different methods.}\label{speed}
		\setlength{\tabcolsep}{0.2cm}
		\resizebox{\linewidth}{!}{
			\begin{tabular}{c|ccc}
				\hline
				Methods & Params (M) & GFLOPs  & Speed (s) \\ \hline
				KernelGAN\cite{kernel_gan}+ZSSR~\cite{zssr}
				&$0.3$ &/ &$415.7$ \\	        
				IKC ~\cite{ikc}   
				& $5.29$       & $2178.72$ & $3.93$      \\
				DANv1~\cite{dan}
				& $4.33$       & $926.72$  & $0.75$      \\ 
				DANv2
				& $4.71$      & $918.12$  & $0.55$      \\  \hline
		\end{tabular}}
	\end{table}
	
	\subsubsection{Inference Speed}
	
	Compared with other blind SR methods, our end-to-end model also has superiority in inference speed. To make a quantitative comparison, we evaluate the average speed of different methods on the same platform. We choose the 40 images synthesized by \textit{Gaussian8}  kernels from Set5 as testing images, and all methods are evaluated on the same platform with an RTX2080Ti GPU. We choose KernelGAN~\cite{kernel_gan} + ZSSR~\cite{zssr} and IKC~\cite{ikc} as the comparison methods. The model complexities and inference speed are shown in Table~\ref{speed}. The FLOPs of KernelGAN+ZSSR is left out because it re-trains a different model for each test image. In that case, FLOPs can not indicate the model complexity. As shown in Table~\ref{speed}, the average speed of DANv1 is \textbf{0.75} seconds per image,  nearly 554 times faster than KernelGAN + ZSSR, and 5 times faster than IKC.  It indicates that DAN not only can largely outperform SOTA blind SR methods on PSNR results but also has a much higher speed. DANv2 further improves the speed of DANv1 by $28\%$.  This is mainly because DPCB removes the expansion and concatenation operation in CRB. It saves many computations and memory and thus can be accelerated.
	
	\subsection{Experiments on Real World Images}
	
	We also conduct experiments to prove that DAN can generalize well to real-world images. We use the model trained with Setting 1 for scale $\times4$ to upscale the commonly used real image  \textit{chip}~\cite{chip}. We use KernelGAN~\cite{kernel_gan} + ZSSR~\cite{zssr} and IKC~\cite{ikc} as the representative methods for blind SR, and CARN~\cite{carn} as the representative method for non-blind SR method. It should be noted that it is a real image and we do not have the ground truth. Thus we can only provide a visual comparison in Figure~\cite{chip}. As one can see, the result of KernelGAN + ZSSR is slightly better than bicubic interpolation but is still heavily blurred. The result of CARN is over smoothed and the edge is not sharp enough. IKC produces a cleaner result, but there are still some artifacts. The letter `X' restored by IKC has an obvious dark line at the top right part. But this dark line is much lighter in the image restored by DAN. It suggests that even if DAN is trained via synthesized image pairs, it still has the ability to generalize to images in real applications in some cases. 
	
	\section{Conclusion}
	
	In this paper, we have proposed an end-to-end algorithm for blind SR. This algorithm is based on alternating optimization, the two parts of which are both implemented by convolutional modules, namely \textit{Restorer} and \textit{Estimator}. We unfold the alternating process to form an end-to-end trainable network. In this way, \textit{Estimator} can utilize information from both LR and SR images, which makes it easier to estimate blur kernel. More importantly, \textit{Restorer} is trained with the kernel estimated by \textit{Estimator}, instead of the ground-truth kernel, thus \textit{Restorer} could be more tolerant to with the estimation error of \textit{Estimator}. Experiments show that the compatibility of the two modules may be more important than their accuracy, and that is the main reason why the proposed method is better than the previous two-step solution. Our main contributions are that we provide an end-to-end algorithm for blind SR and demonstrate that an end-to-end pipeline is important for the final performance. In the future, we will try to apply similar ideas in other low-level vision tasks, such as deblur and denoise. 
	
	\appendices
	\newpage
	{
		\bibliographystyle{IEEEtran}
		\bibliography{egbib}
	}
	
	\begin{IEEEbiography}[{\includegraphics[width=1in]{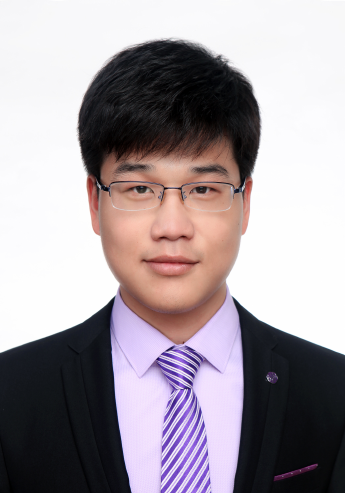}}]{Zhengxiong Luo}
		received the BEng degree in Shanghai Jiao Tong University  (SJTU) in 2018. He is currently a Ph.D. student in the Center for Research on Intelligent Perception and Computing (CRIPAC), National Laboratory of Pattern Recognition (NLPR), Institute of Automation, Chinese Academy of Sciences (CASIA). His current research interests include computer vision and pattern recognition.
	\end{IEEEbiography}
	
	\begin{IEEEbiography}[{\includegraphics[width=1in]{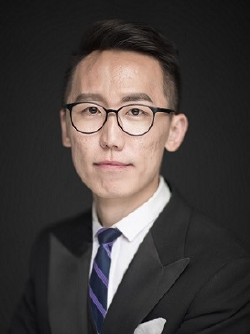}}]{Yan Huang}
		received the BSc degree from the University of Electronic Science and Technology of China (UESTC) in 2012, and the Ph.D. degree from the University of Chinese Academy of Sciences (UCAS) in 2017. He is currently an associate professor working in the National Laboratory of Pattern Recognition (NLPR), Institute of Automation, Chinese Academy of Sciences (CASIA). His research interests include machine learning and pattern recognition. He has published papers in the leading international journals and conferences such as IEEE TPAMI, IEEE TIP, NIPS, ICCV, and CVPR.
	\end{IEEEbiography}
	
	\begin{IEEEbiography}[{\includegraphics[width=1in]{bio/ls.pdf}}]{Shang Li}
		received the B.Eng. degree in University of Electronic Science and Technology of China (UESTC) in 2018. She is currently a Ph.D. student in the Digital Content Technology and Media Service Research Center, Institute of Automation, Chinese Academy of Sciences (CASIA), Beijing, China. Her current research interests include computer vision and pattern recognition.
	\end{IEEEbiography}
	
	\begin{IEEEbiography}[{\includegraphics[width=1in]{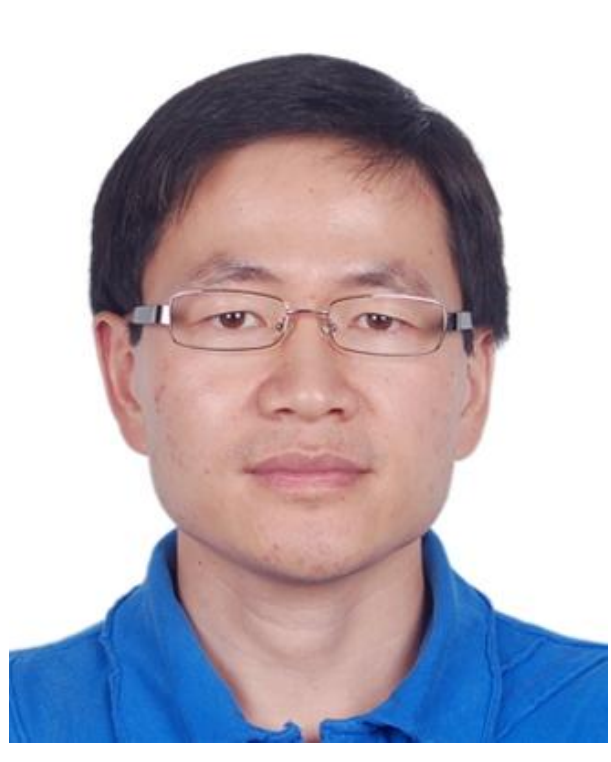}}]{Liang Wang}
		received both the BEng and MEng degrees in Anhui University in 1997 and 2000, respectively, and the Ph.D. degree in the Institute of Automation, Chinese Academy of Sciences (CASIA) in 2004. From 2004 to 2010, he was a research assistant at Imperial College London, United Kingdom, and Monash University, Australia, a research fellow at the University of Melbourne, Australia, and a lecturer at the University of Bath, United Kingdom, respectively. Currently, he is a full professor of the Hundred Talents Program at the National Lab of Pattern Recognition (NLPR), CASIA. His major research interests include machine learning, pattern recognition, and computer vision. He has widely published in highly ranked international journals such as IEEE Transactions on Pattern Analysis and Machine Intelligence and IEEE Transactions on Image Processing, and leading international conferences such as CVPR, ICCV, and ICDM. He is an IEEE Fellow and an IAPR Fellow.
	\end{IEEEbiography}
	
	\begin{IEEEbiography}[{\includegraphics[width=1in]{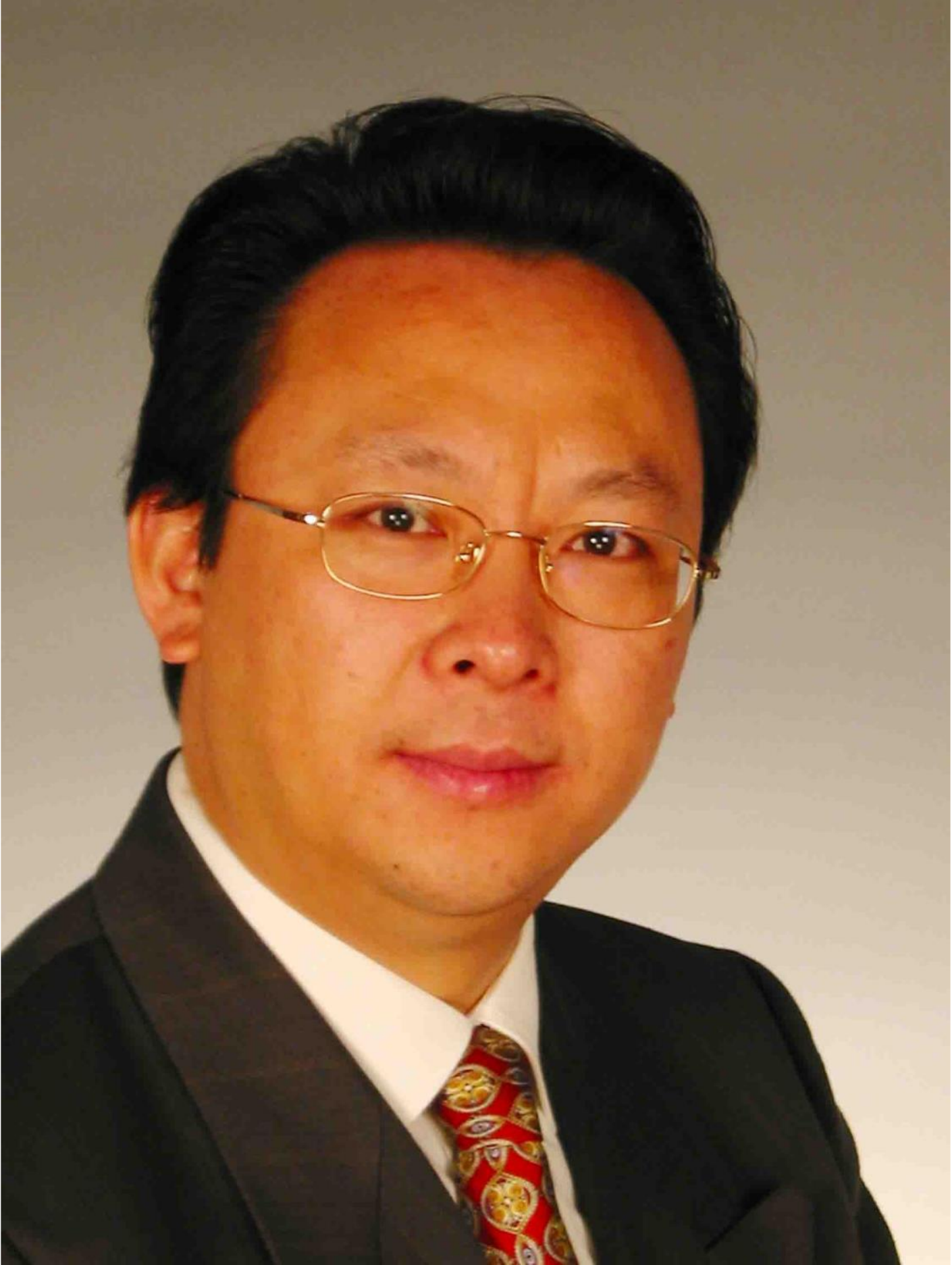}}]{Tieniu Tan}
		received his BSc degree in electronic engineering from Xi'an Jiaotong University, China, in 1984, and his MSc and Ph.D. degrees in electronic engineering from Imperial College London, U.K., in 1986 and 1989, respectively. He is currently a Professor with the Center for Research on Intelligent Perception and Computing (CRIPAC), National Laboratory of Pattern Recognition (NLPR), Institute of Automation, Chinese Academy of Sciences (CASIA). His current research interests include biometrics, image and video understanding, and information forensics and security. He is an IEEE Fellow and an IAPR Fellow.
	\end{IEEEbiography}
	\vfill
	
	
	
	

\end{document}